# MULTIPLE INSTANCE DENSE CONNECTED CONVOLUTION NEURAL NETWORK FOR AERIAL IMAGE SCENE CLASSIFICATION

*Qi Bi [1], Kun Qin[1]\*, Zhili Li[2], Han Zhang[1], Kai Xu[2]*

1. School of Remote Sensing and Information Engineering, Wuhan University, Wuhan, China
2. Faculty of Information Engineering, China University of Geosciences, Wuhan, China
\*Corresponding author, e-mail: qink@whu.edu.cn

**ABSTRACT**

With the development of deep learning, many state-of-the-art natural image scene classification methods have demonstrated impressive performance. While the current convolution neural network tends to extract global features and global semantic information in a scene, the geo-spatial objects can be located at anywhere in an aerial image scene and their spatial arrangement tends to be more complicated. One possible solution is to preserve more local semantic information and enhance feature propagation. In this paper, an end to end multiple instance dense connected convolution neural network (MIDCCNN) is proposed for aerial image scene classification. First, a 23 layer dense connected convolution neural network (DCCNN) is built and served as a backbone to extract convolution features. It is capable of preserving middle and low level convolution features. Then, an attention based multiple instance pooling is proposed to highlight the local semantics in an aerial image scene. Finally, we minimize the loss between the bag-level predictions and the ground truth labels so that the whole framework can be trained directly. Experiments on three aerial image datasets demonstrate that our proposed methods can outperform current baselines by a large margin.

## 1. INTRODUCTION

Earth vision, also known as earth observation and remote sensing, is an important field of computer vision and image understanding [1,2]. Aerial image scene classification, also known as remote sensing image scene classification, is a basic task in Earth vision and aerial image interpretation and is important for a series of aerial image applications, such as land use and land cover (LULC) classification and urban planning. Due to the highly complex geometric structures and spatial patterns in aerial images [3], how to effectively identify the semantic label of an aerial image scene is still a challenging problem for remote sensing community.

Before the development of deep learning, low-level feature based methods [4] and middle-level feature based methods [5–7] are widely used in aerial image scene classification. However, due to the intra-class similarity, inter-class variety and complex spatial arrangement in aerial image, these low and middle level feature based methods lack flexibility and adaptiveness to different scenes.

Since 2012, deep learning methods, also known as the high-level feature based methods, have largely outperformed the aforementioned low-level and middle-level feature based methods in a series of visual tasks including aerial image scene classification [2,8]. Among these deep learning methods, CNN is widely used. Its effectiveness can be explained by its multiple stage feature extraction and its end-to-end framework. Deep learning methods for aerial image scene classification can be mainly divided into three categories, that is, fine tuning, using CNNs as a feature extractor, and full training on aerial image datasets [9,10].

Although these strategies can outperform the traditional hand-crafted feature based methods significantly, several problems still remain. The first problem is that since geo-spatial objects can be located at any corner and at any orientation in aerial image, high-level feature based methods should highlight the local semantics relevant to the scene label [11]. However, current CNNs mainly focus on global semantic features, since in natural image scenes, objects are concentrated and have a more stable context relationship. The second problem is that, gradient vanishing and the loss of middle and low level features can hardly be avoided when training large CNNs [2], while the high variation of texture and structure in aerial image means such low and middle level features need to be well preserved and more discriminative features need to be extracted [12].

The recently proposed DenseNet can preserve the middle and low level features and can alleviate gradient vanishing problems because of its dense connection structure [13]. The basic idea of dense connection is that feature maps of all the preceding layers are used as inputs and its own feature maps are used as inputs into all subsequent layers (demonstrated in Figure.1). Also, dense connection structure is more capable of feature reuse and it enables us to build a deeper CNN without severe over-fitting problems. Meanwhile, with the current boost of deep learning in a series of pattern recognition tasks, multiple instance learning (MIL) has the trend to be combined with CNN to refine semantic labels. MIL is a weak supervised learning in which each of the training samples is regarded as a bag and each bag consists of a series of instances. The key to implement MIL in the visual task is to regard the densely sampled image patches as instances and the scene label is determined by the relevant instances [14,15].

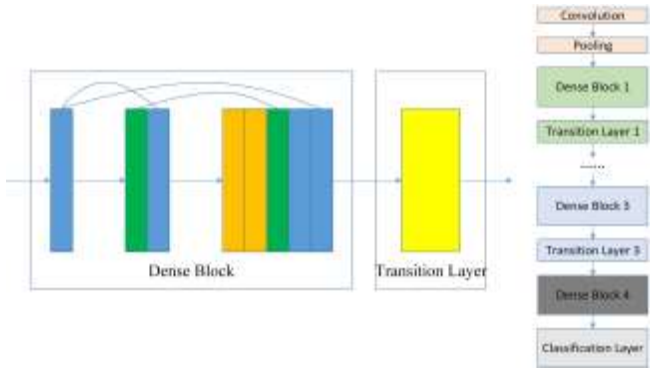

**Fig. 1:** Illustration of dense connected structure. Note that in the left subfigure, each rectangular in this figure stands for a tensor outputted from a convolution layer and in the right subfigure a DenseNet architecture is presented.

Inspired by these studies, we propose a new approach for aerial image scene classification. We list our work and contribution as follows.

(1) We propose a multiple instance dense connected convolutional neural network (MIDCCNN) for aerial image scene classification.

(2) We build a 23-layer dense connected convolutional neural network (DCCNN) served as the backbone in our framework.

(3) We propose an attention-based MIL pooling to highlight the relevant local semantics and it can be directly optimized along with the CNN structure under the supervision of bag label.

## 2. METHODOLOGY
### 2.1. Dense Connected Convolution Neural Network

Let $h_\ell$ denote a composite function of operations and $x_l, x_{l-1}, x_{l-2} \cdots, x_1$ and $x_1$ denote the output of the $l^{th}, (l-1)^{th}, (l-2)^{th}, \cdots, 1^{th}$ and $0^{th}$ layer respectively. The dense connection can be expressed as Equation (1).

$$x_l = H_l([x_0, x_1, \cdots, x_{l-1}]) \quad (1)$$

In DenseNet, the first three dense blocks are followed by a transition layer and the fourth dense block is followed by the classification layers, as is demonstrated in Figure 1. In each dense block, there are a series of 1×1 convolution layer followed by 3×3 convolution layer. In each transition layer, the channel of 1×1 convolution layer are usually less than the channel of inputted feature maps in the hope of feature reduction. However, the DenseNet has hundreds of layers and is too large to train on small aerial image datasets.

**Table 1:** Network structure of DenseNet and the proposed DCCNN

| Layers | Output Size | DenseNet121 | DCCNN |
|---|---|---|---|
| Convolution | 112×112 | 7×7 conv, stride 2 | |
| Pooling | 56×56 | 3×3 max pool, stride 2 | |
| Dense Block 1 | 56×56 | $\begin{bmatrix}1\times1\text{ conv}\\3\times3\text{ conv}\end{bmatrix}\times6$ | $\begin{bmatrix}1\times1\text{ conv}\\3\times3\text{ conv}\end{bmatrix}\times3$ |
| Transition Layer 1 | 56×56 | 1×1 conv | |
| | 28×28 | 2×2 average pool, stride 2 | |
| Dense Block 2 | 28×28 | $\begin{bmatrix}1\times1\text{ conv}\\3\times3\text{ conv}\end{bmatrix}\times12$ | $\begin{bmatrix}1\times1\text{ conv}\\3\times3\text{ conv}\end{bmatrix}\times3$ |
| Transition Layer 2 | 28×28 | 1×1 conv | |
| | 14×14 | 2×2 average pool, stride 2 | |
| Dense Block 3 | 14×14 | $\begin{bmatrix}1\times1\text{ conv}\\3\times3\text{ conv}\end{bmatrix}\times24$ | $\begin{bmatrix}1\times1\text{ conv}\\3\times3\text{ conv}\end{bmatrix}\times3$ |
| Transition Layer 3 | 14×14 | 1×1 conv | |
| | 7×7 | 2×2 average pool, stride 2 | |
| Dense Block 4 | 7×7 | $\begin{bmatrix}1\times1\text{ conv}\\3\times3\text{ conv}\end{bmatrix}\times16$ | None |
| Convolution | 7×7 | None | 1×1 conv |
| Classification Layer | 1×1 | 7×7 global average pool | |
| | | Fully connected, softmax | |

In this paper, we build a 23-layer dense connected convolutional neural network (DCCNN) as a backbone. The difference between our DCCNN and the original DenseNet mainly lies in three aspects:

(1) In each dense block, the number of 3×3 convolutional layers are significantly reduced.

(2) The transition layers serve as the function of feature refinement rather than feature reduction. It is realized by the setting that the number of convolution filters in each transition layer is equal to the number of input feature maps.

(3) The fourth dense block is removed, and instead we add an additional 1×1 convolution layer to further refine the extracted convolution features before classification.

The network structure of DCCNN and DenseNet121 is demonstrated in Table 1 for comparison.

### 2.2. Multiple Instance Learning Pooling

To adapt CNN to multiple instance scenarios, the key idea is to design a MIL pooling layer, which aggregates instance feature vectors into a bag feature vector [16,17]. In this paper, an attention based MIL pooling is proposed to transform CNN into MIL framework and it allows the whole model to be trained directly under the supervision of scene labels (bag labels).

Before MIL pooling, we need to build an instance level classifier for local scene patches. In a CNN, we can obtain a downscaled multi-channel feature map $F$ through multiple convolution layers and pooling layers. Later, the activation $F_{ij}$ of each position $(i, j)$ is computed through convolution and each activation $F_{ij}$ forms a feature vector to represent a local patch. Since instance-level classifier outputs the class predictions $p_{ij}$ of local patches, the dimension of $F_{ij}$ need to match with the number of scene categories.

$$a_{ij} = softmax\left(w_2^T \tanh\left(W_1 F_{ij}^T + b\right)\right) \quad (4)$$

**Table 2**: Detailed Information on Three Aerial Image Datasets

|  | Category | Samples per class | Image size | Training ratio | Spatial resolution |
|---|---|---|---|---|---|
| UCM | 21 | 100 | 256×256 | 80% | 0.3 meter |
| AID | 30 | 220-400 | 600×600 | 50% | 0.5-8 meter |
| NWPU | 45 | 700 | 256×256 | 20% | 0.2-30 meter |

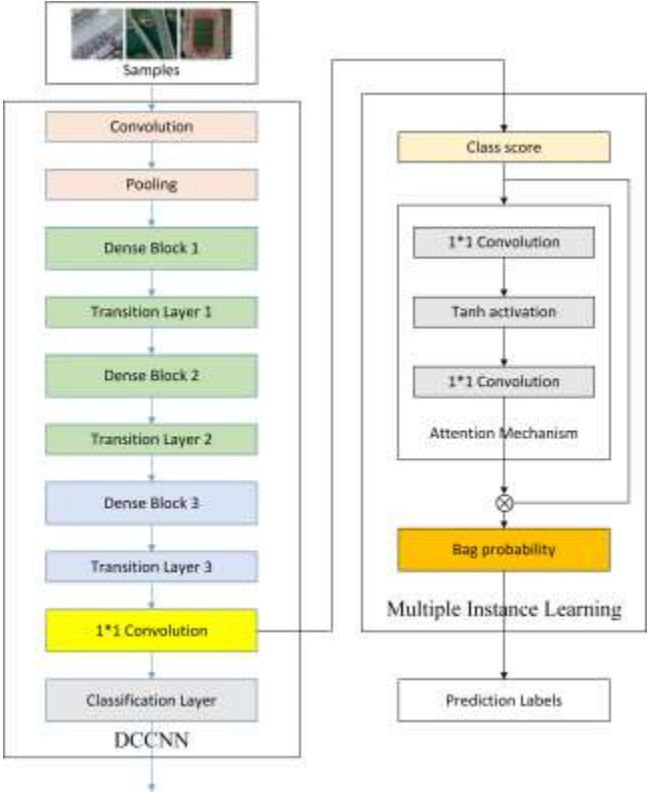

**Fig. 2:** Framework of the proposed MIDCCNN

Since the only accessible supervision information is bag-level labels, a MIL pooling function $g(\cdot)$ is needed to aggregate multiple instance predictions $\{p_{ij}\}$ into a single bag prediction $p_{bag}$, that is:

$$p_{bag} = g\left(\{p_{ij}\}\right) \quad (2)$$

Common approaches to introduce MIL into a CNN framework include using max pooling and average pooling [17]. However, such approaches can not take the complicated spatial arrangement and object distribution in an aerial image into account. To handle this challenge, we propose an attention based MIL pooling.

The spatial attention mechanism [17,19] can select the most relevant local semantics by using a weighted average. Let $\{a_{ij}\}$ denote the attention weights of instances and the sum of $\{a_{ij}\}$ is 1, then the MIL pooling function $g(\cdot)$ can be represented as a convex combination of all instance probability distribution vectors:

$$g\left(\{p_{ij}\}\right) = \sum_i \sum_j a_{ij} p_{ij} \quad (3)$$

In our attention model, we calculate the weights by an attention network, as is demonstrated in Figure 2 and in formula (4). It takes an instance feature vector $F_{ij}$ as the input and it outputs an attention weight $a_{ij}$:

where $w_2 \in \mathbb{R}^L$ and $W_1 \in \mathbb{R}^{L \times N_c}$ are trainable weight parameter matrices of the two layers in the attention network, and $b \in \mathbb{R}^L$ is the bias parameter matrix. We use a softmax function to make the attention weights sum to one.

If the scene contains more than one semantics, this MIL pooling can highlight the instances of interest while suppressing other instances. Otherwise, the weights of all the instances will be roughly equal, and the effect of the attention model will be similar to a global average pooling.

### 2.3. MIDCCNN Framework

The framework of the proposed multiple instance dense connected convolution neural network (MIDCCNN) is demonstrated in Figure 2. In this framework, at first we use DCCNN as a backbone to extract the convolution features. We feed the convolution features outputted from the additional 1x1 convolution layer into the instance classifier. The instance classifier consists of N 1x1 convolution filters (Here N is equal to the number of scene categories) to compute the instance-level feature vectors. Then, we use the proposed MIL pooling to obtain a bag-level class probability. Finally, we use the classic cross entropy loss function to minimize the loss between the bag-level predictions and the ground truth labels, and the whole framework can be optimized as a whole. It should be noted that the major difference between our MIL pooling and the attention based MIL pooling proposed in [20] is that in [21] it can not be supervised directly under the bag label and thus lacks the interpretability of a result.

### 3. EXPERIMENT AND ANALYSIS
#### 3.1. Dataset
We implement our experiments on three widely used aerial image scene classification datasets, that is, the UCM dataset [21], the AID dataset [2] and the NWPU dataset [8]. The basic information of these three datasets and their corresponding training ratios are listed in Table 2.

#### 3.2. Model initialization and training
For the parameter initialization of our DCCNN and MIDCCNN, we use random initialization for weight parameters and all bias parameters are set to 0.001, respectively.

The model was trained by the Adam optimizer, with a two-stage training strategy. On AID and NPWU dataset, the learning rate was set to 0.001 for 100 epochs at the first stage. After 100 epochs, the learning rate was divided by 10 and we continue training until the termination. On UCM dataset, the learning rate was set to 0.001 for 40 epochs at the first stage. After 40 epochs, the learning rate was divided by 10 and we continue training until the termination. The training process does not terminate until the learning rate drops to 1e-6. During the whole training progress, to prevent over fitting problem, L2 normalization is used to with a parameter setting of 1e-6 and dropout rate is set to be 0.2.

Table 3: Overall accuracy (OA) on three aerial image datasets

|  | UCM | AID | NWPU |
|---|---|---|---|
| SIFT [2] | 32.10±1.95 | 16.76±0.65 | - |
| PLSA [2] | 71.38±1.77 | 63.07±0.48 | - |
| BoVW [2,10] | 75.52±2.13 | 68.37±0.40 | 44.97±0.28 |
| LDA [2] | 75.98±1.60 | 68.96±0.58 | - |
| AlexNet [2,10] | 95.02±0.81 | 89.53±0.31 | 79.85±0.13 |
| VGGNet [2,10] | 95.21±1.20 | 89.64±0.36 | 79.79±0.15 |
| GoogLeNet[2,10] | 94.31±0.89 | 86.39±0.55 | 78.48±0.26 |
| MIL_mean [17,20] | 96.41±0.44 | 92.19±0.24 | 86.37±0.18 |
| MIL_max [17,20] | 95.91±0.55 | 91.21±0.27 | 85.23±0.21 |
| DCCNN (ours) | **96.21±0.67** | **91.49±0.22** | **85.63±0.18** |
| MIDCCNN (ours) | **97.00±0.49** | **92.53±0.18** | **87.32±0.17** |

Meanwhile, in accordance with number of scene categories, the number of convolution filters in the instance classifier on UCM, AID and NPWU dataset is set to be 21, 30 and 45, respectively.

### 3.3. Results and discussion

The experimental evaluation follows the widely accepted evaluation protocol of aerial image scene classification [2]. To be specific, we compute the overall accuracy (OA) for ten repetitions and then compare its average and standard deviation with other baseline methods reported in [2] and [10]. Meanwhile, to validate the effectiveness of our proposed attention-based MIL pooling, while keeping the backbone (DCCNN) the same, we also compare our method with currently used maximum and average pooling operations in MIL [17,20] (denoted as MIL_mean and MIL_max respectively). All results are listed in Table 3.

From these experimental results, some important phenomenon can be observed.

(1) Our proposed DCCNN and MIDCCNN achieves the highest OA on all three aerial image datasets and can outperform other baseline deep learning models. Since there are relatively low number of samples in the UCM dataset, the improvement of OA is not significant. While there are a relatively larger amount of samples in AID and NWPU dataset, the middle and low level feature methods perform much worse and the boost of OA is significant when using our proposed method. It can be explained by the utilization of dense connection and our attention based MIL pooling. The dense connection structure is more capable of feature propagation while having much fewer parameters and the fact that our attention based MIL pooling help highlight the important local semantics in an aerial scene.

(2) When using the same convolution feature extraction

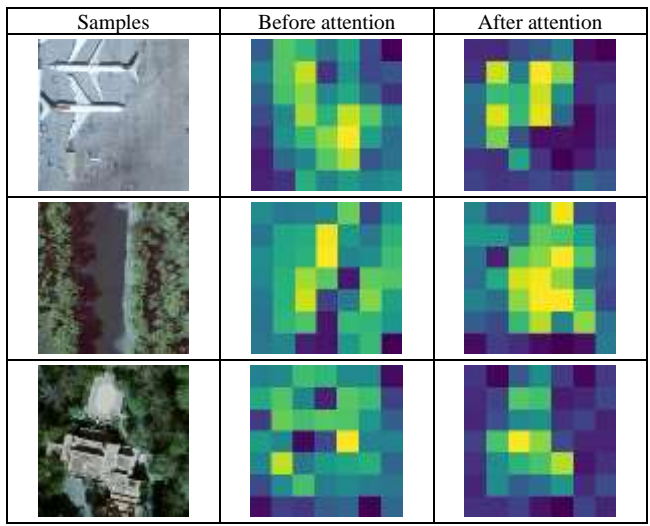

Fig. 3: Demonstration of attention based MIL pooling

framework, our proposed attention based MIL pooling outperforms other currently used MIL pooling methods. It might be explained by the following two aspects. The first aspect is that compared with common used maximum and average pooling operations, attention based pooling operations are capable of highlighting the instances relevant to the bag label and at the same time suppressing other instances by assigning different weights. Feature maps of several samples before and after the attention pooling help illustrate this (Seen in Figure 3). The second aspect is that compared with other MIL pooling operations, our method is directly under the supervision of bag-level labels and thus has a better scene feature representation ability.

### 4. CONCLUSION

In this paper, we propose a multiple instance dense connected convolution neural network (MIDCCNN) for aerial image scene classification. The backbone is a 23-layer dense connected convolution neural network (DCCNN) and its effectiveness mainly lies in dense connection, little amount of parameters and a series of 1×1 convolutional layers. To implement MIL under a CNN architecture, we propose a MIL pooling based on spatial attention mechanism and it can be optimized together with CNN under the bag-level supervision. Experiments on three aerial image datasets demonstrate that the proposed method can outperform other state-of-the-art methods since it can extract more discriminative features well and highlight the local semantics in an aerial image scene.

Future works include studying feature reuse in our MIDCCNN and introducing our framework to instance segmentation and object detection in aerial image.

## 6. ACKNOWLEDGEMENT

This research is supported by the National Key Research and Development Program of China (No. 2016YFB0502600).